\pdfoutput=1
\documentclass[letterpaper, 10 pt, conference]{ieeeconf}

\IEEEoverridecommandlockouts

\usepackage{amsmath}
\usepackage{amssymb}
\usepackage{booktabs}
\usepackage{graphicx}
\usepackage{placeins}
\usepackage{xcolor}
\usepackage{tikz}
\usepackage{pgfplots}
\pgfplotsset{compat=1.18}
\usetikzlibrary{arrows.meta,calc,positioning}

\usepackage{eso-pic}
\AddToShipoutPictureBG*{%
  \AtPageUpperLeft{%
    \raisebox{-0.5in}{\makebox[\paperwidth][c]{%
      \parbox{0.82\paperwidth}{\centering\footnotesize
        This work has been submitted to the IEEE for possible publication.
        Copyright may be transferred without notice, after which this
        version may no longer be accessible.}}}}}

\usepackage[hidelinks]{hyperref}
\hypersetup{
  pdftitle={ViewPACT: Multi-View Geometric Augmentation and Pose-Action Co-Training for Viewpoint-Robust Manipulation},
  pdfauthor={Chenglang Tu, Feng Chen, Ruizhe Liu, Pei Zhou, Binbin Huang, Chen Xu, Yanchao Yang, Yi Ma, Shenghua Gao},
}

\definecolor{vpblue}{HTML}{3B6FB6}
\definecolor{vporange}{HTML}{D9772B}
\definecolor{vpgreen}{HTML}{2F8A70}
\definecolor{vpgray}{HTML}{5F6772}

\newcommand{\method}{\textsc{ViewPACT}}

\title{\LARGE \bf
\method: Multi-View Geometric Augmentation and Pose--Action Co-Training for
Viewpoint-Robust Manipulation
}

\author{\normalsize Chenglang Tu$^{1,*,\ddagger}$, Feng Chen$^{2,*}$, Ruizhe Liu$^{2}$, Pei Zhou$^{2}$, Binbin Huang$^{2}$,\\ Chen Xu$^{2}$, Yanchao Yang$^{2,\dagger}$, Yi Ma$^{2,\dagger}$, Shenghua Gao$^{2,\dagger}$%
\thanks{$^{*}$Equal contribution. $^{\dagger}$Corresponding authors. $^{\ddagger}$Work done during an internship at The University of Hong Kong.}%
\thanks{$^{1}$University of Science and Technology of China. $^{2}$The University of Hong Kong.}}

\begin{document}

\maketitle
\thispagestyle{empty}
\pagestyle{empty}

\begin{abstract}
Image policies are scalable but remain brittle to camera viewpoint changes. On LIBERO, we identify proprioception as an input shortcut: masking robot states sharply degrades several policies, while retraining from images alone recovers much of the loss. We introduce \method{}, an image policy trained with privileged geometric supervision. Auxiliary objectives recover end-effector and task-relevant object poses in the camera frame, while deployment requires only RGB observations. Joint geometric supervision improves mean simulation success from $37.4\%$ to $78.4\%$ under unseen external-camera viewpoints, with more modest gains under separately evaluated object, scene, and camera-intrinsic shifts. Across four real-robot tasks, combining simulated and real-image geometric supervision achieves $68.1\%$ mean success, versus $3.1\%$ for fixed-view behavior cloning and $30.6\%$ for the strongest evaluated alternative. These results support geometric supervision for viewpoint robustness in the evaluated fixed-view demonstration setting.
\end{abstract}

\section{Introduction}

Image-based policies are attractive because RGB observations are inexpensive and abundant in robot demonstrations and large-scale video. Yet they often generalize less effectively to unseen camera viewpoints than policies using explicit 3D representations
\cite{ze2025generalizablehumanoidmanipulation3d, ze2024dp3}.
While 3D representations encode robot--object configurations in geometrically stable coordinates, image appearance varies with camera pose and intrinsics. The challenge is to introduce geometric knowledge without requiring 3D observations or reconstruction at deployment.

Existing approaches transfer pretrained geometric features through
feature fusion or representation alignment
\cite{yu20263dmixvlaplugandplaymodule,li2025spatialforcingimplicitspatial},
or broaden training coverage through demonstration generation
\cite{mandlekar2023mimicgen,xue2025demogen}, novel-view synthesis
\cite{tian2025vista}, and scene reconstruction
\cite{yang2025robosplat}. These approaches introduce geometry through
pretrained representations or augmented observation--action data.
We investigate a complementary transfer strategy: using simulation
to supervise task-relevant robot--object geometry rather than
synthetic actions, while retaining real demonstrations as the
source of action supervision.

We find that image-based policies can develop strong dependence on
proprioceptive inputs~\cite{lu2026visionproprioceptionpoliciesfailrobotic}
even when training without those inputs achieves
higher task success. On LIBERO~\cite{liu2023libero}
(Sec.~\ref{sec:prop_analysis}), test-time
state masking substantially degrades several state-conditioned policies,
whereas retraining without the ablated inputs recovers substantial
performance. This contrast demonstrates that
strong learned input dependence does not establish that the input is
necessary to attain the corresponding control performance. These
findings motivate using robot-state information to supervise visual
representations rather than directly condition action prediction.
Since auxiliary robot-state prediction alone does not consistently
improve control, we introduce privileged geometric supervision of
both end-effector and object poses during training, explicitly
targeting robot--object geometry while retaining image-only action
prediction at inference.

We introduce \method{}, which uses end-effector and task-relevant object poses as privileged training targets rather than policy inputs. Actions are predicted solely from images, while auxiliary objectives require the visual representation to recover these poses in the camera frame, encouraging it to encode robot--object configurations and spatial relationships. Deployment requires only visual observations, without pose labels, proprioception, or simulator states.

We evaluate \method{} in controlled simulation and on four real-robot tasks. In simulation, we isolate joint pose supervision and interaction-aware state sampling and test robustness to viewpoint, object, scene, and camera-intrinsic shifts. Joint end-effector and object supervision raises mean success from $37.4\%$ to $78.4\%$. On the real robot, we compare simulated geometric supervision with simulated action augmentation and novel-view synthesis, and examine zero-shot real-image pose augmentation as a complementary source. Combining simulated and real-image geometric supervision achieves $68.1\%$ mean success, versus $3.1\%$ for fixed-view behavior cloning and $30.6\%$ for the strongest alternative. These results show that, in the studied low-cost fixed-view regime, simulation transfers more effectively through geometric than action supervision, while simulated and real-image geometric supervision provide complementary benefits.

\begin{figure*}[!t]
  \centering
  \includegraphics[width=\textwidth]{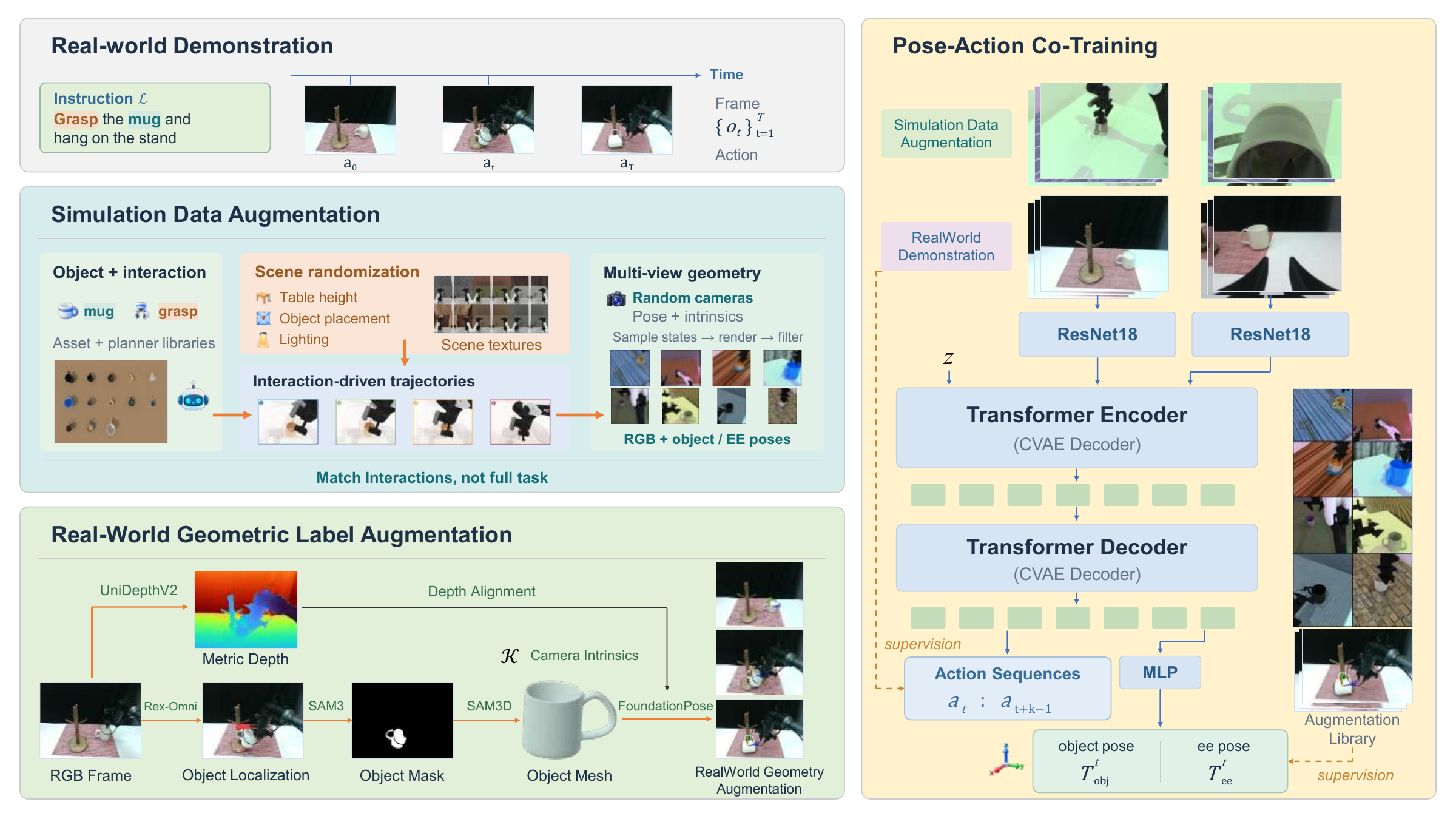}
  \caption{\normalfont Overview of \method{}. Starting from fixed-view real-world demonstrations, the framework
  augments supervision with simulation-based multi-view geometric data and zero-shot real-image pose annotations.
  The resulting pose--action co-training objective encourages the visual representation to encode task-relevant
  object and end-effector geometry while requiring only RGB observations at deployment.}
  \label{fig:method_overview}
\end{figure*}

\section{RELATED WORK}
\label{sec:related_work}

\subsection{Data Augmentation for Robot Learning}
Simulation-based augmentation expands action-labeled data through
trajectory retargeting, skill composition, and joint transformation
of actions and 3D observations
\cite{mandlekar2023mimicgen,garrett2025skillmimicgen,xue2025demogen}.
These methods scale action supervision but typically rely on
task-specific structure.

Generative editing broadens appearance coverage
\cite{chen2023genaug,chen2025roviaug}, while novel-view
synthesis augments camera viewpoints
\cite{tian2025vista,cai2026robonvs}. Segmentation and generation errors
can nevertheless distort task-relevant geometry, particularly under
large viewpoint changes. Explicit reconstruction supports interactive
real-to-sim replicas \cite{torne2024rialto,barcellona2025drema} and
view-consistent 3D Gaussian splatting (3DGS) augmentation
\cite{yang2025robosplat}, but introduces scene-specific
modeling, calibration, or multi-view capture requirements.
In contrast, \method{} requires neither deployment-scene reconstruction
nor additional real-world data collection.

\subsection{Geometry-Grounded Learning for Manipulation}
Geometry-grounded policies represent depth or 3D observations as
voxels, point clouds, virtual views, or end-effector-centric features
\cite{shridhar2023peract,goyal2023rvt,ze2024dp3,wilcox2025adapt3r,yuan2025maniwhere}.
These representations promote viewpoint robustness, but their
deployment-time depth requirements introduce sensing overhead and
sensitivity to measurement noise.

Another line transfers geometric priors through feature distillation
or geometric-backbone fine-tuning
\cite{shen2023f3rm,guo2025glad,han2026gam}, with additional computational
cost. Other methods learn action-relevant representations, including
semantic--geometric features, keypoints, view-consistent features,
motion latents, and camera-frame actions or \(\mathrm{SE}(3)\)
trajectories
\cite{zhang2023sgr,vecerik2021s3k,yuan2025maniwhere,li2026camvla,chen2026oasis}.
Learning these representations often requires multi-view or RGB-D
observations, and some methods additionally rely on manual geometric
annotations. \method{} instead learns from simulated pose labels and
automatically annotated fixed-view real images. At deployment, it
uses only RGB observations, without pose annotations or 3D sensing.

\subsection{Sim--Real Co-Training for Robot Manipulation}
Classical sim-to-real methods address the reality gap through visual
randomization or domain adaptation
\cite{tobin2017domain,bousmalis2018simulation}. Recent approaches combine
simulated and real action trajectories or pair simulated robot
experience with real human observations
\cite{maddukuri2025simreal,fang2026simhum}, while others align joint
observation--action distributions across domains
\cite{cheng2025generalizable}. Transferring simulated action supervision
requires addressing discrepancies in dynamics, embodiment, and action
distributions. Our method instead uses simulation to supervise
camera-frame end-effector and object poses, rather than actions.
This provides task-relevant geometric supervision without requiring
simulated actions or dynamics to transfer to the physical system.

\section{Proprioceptive Dependence and Visual Control}
\label{sec:prop_analysis}

We distinguish a trained policy's dependence on proprioceptive inputs
from the benefit of providing those inputs during learning. To examine
this distinction, we compare three treatments of robot-state information:
masking state inputs only at test time, removing them during both
training and evaluation, and replacing them with auxiliary prediction
targets. For ACT~\cite{zhao2023learning}, DP-UNet, and
DP-DiT~\cite{chi2023diffusion}, test-time masking sets the
seven joint-state coordinates and three end-effector-position
coordinates to zero. We report equal-weight means of suite-level
success rates across LIBERO-10, LIBERO-90, LIBERO-Goal,
LIBERO-Spatial, and LIBERO-Object.

\paragraph{Test-time masking reveals learned input dependence}
Table~\ref{tab:prop_mask} shows that masking state inputs sharply
degrades several evaluated policies. ACT decreases from $64.6\%$
to $19.3\%$, while DP-UNet and DP-DiT nearly collapse, dropping
from $65.1\%$ to $1.6\%$ and from $52.7\%$ to $0.9\%$,
respectively. In contrast, the evaluated $\pi_{0.5}$ checkpoint
retains almost identical performance, achieving $96.6\%$ with
state inputs and $96.5\%$ after masking. These results establish
strong state-input dependence in the evaluated ACT and DP
configurations. However, masking changes the inputs of an
already-trained policy; it does not determine the performance
achievable by training without those inputs.

\begin{table}[t]
    \centering
    \caption{
    Test-time state masking on LIBERO.}
    \vspace{-8pt}
    \label{tab:prop_mask}
    \small
    \begin{tabular}{lrrr}
        \toprule
        Model
        & With state
        & Masked
        & Drop \\
        \midrule
        $\pi_{0.5}$~\cite{intelligence2025pi05visionlanguageactionmodelopenworld}
        & 96.6
        & 96.5
        & 0.1 \\
        ACT~\cite{zhao2023learning}
        & 64.6
        & 19.3
        & 45.3 \\
        DP-UNet~\cite{chi2023diffusion}
        & 65.1
        & 1.6
        & 63.5 \\
        DP-DiT~\cite{chi2023diffusion}
        & 52.7
        & 0.9
        & 51.9 \\
        \bottomrule
    \end{tabular}
    \vspace{-8pt}
\end{table}

\paragraph{Retraining separates dependence from input necessity}
When the state inputs are removed throughout training and evaluation,
performance substantially exceeds the test-time-masking results
(Table~\ref{tab:prop_training}). ACT achieves $67.8\%$ success,
surpassing its $64.6\%$ state-conditioned baseline. DP-DiT and
DP-UNet recover to $45.0\%$ and $50.5\%$, respectively, although
both remain below their state-conditioned baselines. Thus, state
inputs improve the achieved performance of these two DP
configurations, but are not required for ACT to reach its
state-conditioned baseline.

The DINO experiments further show that the benefit of state inputs
varies within the diffusion-policy family.
DP-DiT achieves $52.7\%$ success with state inputs and $45.0\%$
without them, whereas DP-DiT-DINO achieves $68.6\%$ and $68.0\%$,
respectively. A large performance penalty from state-input removal
is therefore not shared by all evaluated diffusion-policy
configurations.

\begin{table}[t]
    \centering
    \caption{
    Training-time state-input ablations.
    Mean success (\%).
    }
    \vspace{-8pt}
    \label{tab:prop_training}
    \small
    \setlength{\tabcolsep}{4pt}
    \begin{tabular}{lrrr}
        \toprule
        Model
        & State input
        & Input removed
        & State target \\
        \midrule
        ACT
        & 64.6
        & \textbf{67.8}
        & 65.1 \\
        DP-DiT
        & \textbf{52.7}
        & 45.0
        & 48.7 \\
        DP-UNet
        & \textbf{65.1}
        & 50.5
        & 45.5 \\
        DP-DiT-DINO
        & \textbf{68.6}
        & 68.0
        & 58.2 \\
        \bottomrule
    \end{tabular}
    \vspace{-8pt}
\end{table}

\paragraph{Auxiliary state prediction provides inconsistent gains}
We next use robot state as an auxiliary prediction target rather
than as a policy input. Comparing this treatment against
input removal isolates the contribution of state supervision
(Table~\ref{tab:prop_training}). State prediction improves
DP-DiT from $45.0\%$ to $48.7\%$, but reduces ACT from
$67.8\%$ to $65.1\%$ and DP-UNet from $50.5\%$ to $45.5\%$.
The DINO configuration exhibits the same limitation:
adding state prediction decreases its four-suite mean from
$68.0\%$ to $58.2\%$. Moving robot state from the input to
a prediction target therefore does not consistently improve
control.

These results motivate a distinction between removing a direct
state-input dependency and providing effective supervision for
visual control. Robot-state reconstruction supervises the robot
configuration but does not explicitly supervise the manipulated
objects. Our method instead removes proprioceptive policy inputs
and jointly supervises end-effector and object poses during
training, directing the visual representation toward the
robot--object geometry required for manipulation.

\section{METHOD}
\label{sec:method}

\method{} combines expert action supervision with two sources of
geometric supervision: simulated object and end-effector poses, and
automatically estimated object poses from existing real images.
Geometric labels supervise the visual representation during training
but are not policy inputs. No additional real-world data collection
is required.

\subsection{Problem Formulation}
\label{sec:problem_formulation}

We assume fixed-view expert demonstrations
\begin{equation}
\begin{aligned}
\mathcal D_{\mathrm{act}}
&= \{(o_t^{c_0},A_t)\},
\qquad A_t=a_{t:t+H-1},\\
o_t^c
&= \left(I_t^{\mathrm{ext},c},I_t^{\mathrm{wrist}}\right),
\end{aligned}
\end{equation}
where $c_0$ is the external-camera configuration used for collection
and $H$ is the action-chunk length. Each observation contains external-
and wrist-camera RGB images. The demonstrations contain neither
additional external viewpoints nor explicit 3D annotations. Our goal
is to learn a policy $\pi_\theta$ that predicts action chunks from
RGB observations and generalizes zero-shot to unseen external-camera
configurations $c\in\mathcal C_{\mathrm{OOD}}$.

\subsection{Simulation-Based Geometric Augmentation}
\label{sec:simulation_augmentation}

We use simulation to generate interaction-relevant images and pose
labels, not action targets. Scenes are constructed from category-matched
assets without reconstructing the deployment scene or calibrating the
simulator to the physical setup.

\paragraph{Interaction-state sampling}
Given a task instruction, we identify the target object and interaction
primitive. We randomize the object instance, initial pose, scale,
appearance, and articulated state, together with the background,
illumination, and table height. A primitive-conditioned motion planner
generates feasible robot--object trajectories. We retain successful
trajectories and sample states across their temporal phases for
multi-view rendering.

\paragraph{Multi-view rendering and pose labels}
For each retained state $s_i$, we sample $V$ external cameras around
a task-space center $\mathbf m_i$. Camera positions are parameterized by
\begin{equation}
\mathbf e_{i,v}
=
\mathbf m_i
+
r_{i,v}
\left(
\cos\phi_{i,v},
\sin\phi_{i,v},
\tan\alpha_{i,v}
\right)^\top,
\end{equation}
where $\phi_{i,v}$ is the azimuth, $\alpha_{i,v}$ is the elevation
measured from the horizontal plane, and $r_{i,v}$ is the horizontal
radius. We reject and resample views with inadequate framing,
visibility, or excessive occlusion of the target, robot, or gripper.
Each accepted view contributes
\begin{equation}
d_{i,v}
=
\left(
o_i^{c_{i,v}},
P_{i,v}^{\mathrm{obj}},
P_{i,v}^{\mathrm{ee}}
\right)
\end{equation}
to $\mathcal D_{\mathrm{sim}}^{\mathrm{geo}}$.
Both poses are expressed in the external-camera frame, aligning the
targets with the rendered viewpoint without assuming shared
robot-base coordinates across environments. The wrist image provides
additional context but does not define the supervision frame.

\subsection{Zero-Shot Real-Image Pose Annotation}
\label{sec:real_world_augmentation}

To complement simulated supervision with real observations, we generate
object-pose pseudo-labels for the existing demonstrations. Given the
target object name, REX-Omni and SAM3 localize and segment the object,
and UniDepth predicts metric depth. SAM3D reconstructs a canonical
mesh from a reference view with clear object visibility. This mesh
is reused across episodes to maintain a consistent object frame.

We estimate its metric scale by matching rendered silhouettes and
depth to the reference mask $M$ and predicted depth $D$:
\begin{equation}
\begin{aligned}
s^*=\arg\max_s\;&
\lambda_{\mathrm{mask}}\operatorname{IoU}(M,\hat M_s)\\
&+
\lambda_{\mathrm{depth}}
\exp\!\left(
-\operatorname*{median}_{u\in\Omega_s}
\left|\log\frac{D(u)}{\hat D_s(u)}\right|
\right),
\end{aligned}
\end{equation}
where $\hat M_s$ and $\hat D_s$ are the rendered mask and depth at
scale $s$. The overlap set $\Omega_s$ contains pixels inside both
masks with positive predicted and rendered depth; the depth score
is zero when this set is empty. We search logarithmically spaced
scales in a coarse-to-fine manner. FoundationPose then registers and
tracks the scaled mesh throughout the demonstrations, producing
$\mathcal D_{\mathrm{real}}^{\mathrm{geo}}
=\{(o_t^{c_0},P_t^{\mathrm{obj}})\}$.
These real samples provide object labels only.

\subsection{Pose--Action Co-Training}
\label{sec:pose_action_cotraining}

The policy shares its visual representation between action prediction
and geometric supervision. Two auxiliary decoder queries and a
lightweight prediction head estimate object and end-effector poses.
At each training step, we sample one mini-batch from
$\mathcal D_{\mathrm{act}}$ and one from
$\mathcal D_{\mathrm{geo}}
=\mathcal D_{\mathrm{sim}}^{\mathrm{geo}}
\cup\mathcal D_{\mathrm{real}}^{\mathrm{geo}}$.

Action loss is evaluated only on expert demonstrations. Geometric
loss uses the annotated entities of each sample:
$\mathcal K(d)=\{\mathrm{obj},\mathrm{ee}\}$ for simulated samples and
$\mathcal K(d)=\{\mathrm{obj}\}$ for real samples.
The objectives are
\begin{subequations}\label{eq:training_objective}
\begin{align}
\mathcal L
&=
\mathcal L_{\mathrm{act}}
+\lambda_{\mathrm{geo}}\mathcal L_{\mathrm{geo}},
\label{eq:total_objective}
\\
\mathcal L_{\mathrm{act}}
&=
\mathbb E_{(o_t,A_t)\sim\mathcal D_{\mathrm{act}}}
\biggl[
\frac{1}{|\mathcal V_t|}
\sum_{h\in\mathcal V_t}
\|\hat a_{t+h}-a_{t+h}\|_1
\notag\\
&\qquad
+\lambda_{\mathrm{KL}}
D_{\mathrm{KL}}\!\left(
q_\theta(z\mid o_t,A_t)
\,\middle\|\,
\mathcal N(0,I)
\right)
\biggr],
\label{eq:action_objective}
\\
\mathcal L_{\mathrm{geo}}
&=
\mathbb E_{d\sim\mathcal D_{\mathrm{geo}}}
\left[
\sum_{k\in\mathcal K(d)}
\|\hat P^k(o_d)-P_d^k\|_2^2
\right].
\label{eq:geometry_objective}
\end{align}
\end{subequations}
Here, $\mathcal V_t$ indexes valid, non-padded actions,
$q_\theta$ is the training-time posterior over the latent variable
$z$, and $P^k$ denotes the vector representation of the corresponding
camera-frame pose. At deployment, the policy predicts action chunks
from RGB observations alone, without proprioceptive inputs,
geometric labels, or the annotation pipeline.

\section{SIMULATION EXPERIMENTS}
\label{sec:experiments}

We design our experiments to answer the following questions: (1) Does pose--action co-training improve closed-loop manipulation under unseen camera viewpoints? (2) What are the individual and joint contributions of object-pose and end-effector-pose supervision? (3) Does interaction-aware state sampling improve the geometric data-augmentation pipeline over random-state augmentation? (4) Does the learned representation remain robust under object, scene, and camera-intrinsic distribution shifts?

\begin{table*}[!t]
    \centering
    \caption{\normalfont Unseen-view closed-loop success rates (\%). The baseline uses action supervision only; the other variants add the indicated camera-frame pose supervision on augmented observations. Panel (a) reports the mean across held-out viewpoints. Panel (b) compares each variant with the seed-matched baseline across task--view pairs and reports the worst-view success rate. Best results are bold.}
    \label{tab:unseen_view}
    \small
    \setlength{\tabcolsep}{4pt}
    \begin{tabular*}{\textwidth}{@{\extracolsep{\fill}}lrrrrrr@{}}
        \toprule
        \multicolumn{7}{l}{\textbf{(a) Average success rate by task.}} \\
        \cmidrule(lr){1-7}
        Method & Hanging Mug & \shortstack{Move Playing\\Card} & Open Laptop & \shortstack{Place Empty\\Cup} & \shortstack{Put Object\\in Cabinet} & Average \\
        \midrule
        Baseline & 60 & 11 & 36 & 17 & 63 & 37.4 \\
        + Object Pose & 70 & 47 & 42 & 23 & 94 & 55.2 \\
        + EE Pose & 74 & 24 & 12 & 31 & 85 & 45.2 \\
        \textbf{\method{} (Object + EE Pose)} & \textbf{89} & \textbf{64} & \textbf{89} & \textbf{52} & \textbf{98} & \textbf{78.4} \\
        \midrule
        \multicolumn{7}{l}{\textbf{(b) Comparison with the baseline across task--view pairs.}} \\
        \cmidrule(lr){1-7}
        Method & \multicolumn{3}{c}{Task--View Pairs $\geq$ Baseline} & \multicolumn{3}{c}{Worst View SR} \\
        \midrule
        Baseline & \multicolumn{3}{c}{--} & \multicolumn{3}{c}{0} \\
        + Object Pose & \multicolumn{3}{c}{41 / 50} & \multicolumn{3}{c}{0} \\
        + EE Pose & \multicolumn{3}{c}{40 / 50} & \multicolumn{3}{c}{0} \\
        \textbf{\method{} (Object + EE Pose)} & \multicolumn{3}{c}{\textbf{50 / 50}} & \multicolumn{3}{c}{\textbf{20}} \\
        \bottomrule
    \end{tabular*}
\end{table*}

\subsection{Experimental Setup}

\subsubsection{Task}

We conduct simulation experiments in RoboTwin 2.0~\cite{chen2025robotwin2}, which is built on the SAPIEN physics simulator. We select five tasks: \textbf{Hanging Mug}, \textbf{Move Playing Card Away}, \textbf{Open Laptop}, \textbf{Place Empty Cup}, and \textbf{Put Object in Cabinet}. We adapt the original benchmark configurations for both data collection and evaluation to construct a controlled testbed for viewpoint generalization.

\subsubsection{Data}

For each task, we collect 100 successful motion-planned demonstrations using the ALOHA-AgileX embodiment. Each demonstration contains action labels and RGB observations from a fixed external camera and a wrist camera. Geometric data are generated using the augmentation pipeline described in Sec.~\ref{sec:simulation_augmentation}.

\subsubsection{Evaluation}

All methods use the same Action Chunking Transformer (ACT) backbone. For each task and supervision setting, we select the checkpoint with the lowest validation loss.

\subsection{Generalization to Unseen Camera Viewpoints}

We first evaluate whether geometric data augmentation and pose--action co-training improve generalization to unseen camera viewpoints.

As shown in Table~\ref{tab:unseen_view}(a), supervising either object pose or end-effector pose alone yields task-dependent gains and can even degrade performance; for example, end-effector-pose supervision performs worse than the baseline on Open Laptop. In contrast, joint supervision achieves the best performance across all tasks. A possible explanation is that the object and end effector together capture the core semantics of manipulation: the object specifies what should be acted upon, while the end effector specifies the acting body.

Table~\ref{tab:unseen_view}(b) further shows that the gain from joint supervision is consistent across viewpoints. The single-target variants can still underperform the paired baseline at particular camera configurations, whereas the joint model matches or exceeds it across all evaluated task--view pairs and remains stronger under the most challenging views.

\subsection{Random versus Interaction-Aware Augmentation}

We compare interaction-aware state sampling with both the unaugmented baseline
and a random-state variant. The two augmented variants use the same object
assets, scene and camera distributions, data budget, pose supervision, and
training configuration; they differ only in state selection. Random-state
augmentation independently samples object states and robot joint configurations,
whereas interaction-aware augmentation draws feasible states from manipulation
trajectories.

\begin{figure}[t]
  \centering
  \makebox[\columnwidth][c]{%
    \begin{tikzpicture}
    \begin{axis}[
        width=\columnwidth,
        height=0.69\columnwidth,
        ybar=2pt,
        bar width=7pt,
        enlarge x limits=0.08,
        ymin=0,
        ymax=120,
        ylabel={Success rate (\%)},
        symbolic x coords={Mug, Card, Laptop, Cup, Cabinet, Avg.},
        xtick=data,
        ytick={0,20,40,60,80,100},
        xticklabel style={font=\scriptsize},
        tick label style={font=\scriptsize},
        label style={font=\small},
        axis line style={vpgray!65},
        tick style={vpgray!65},
        ymajorgrids=true,
        major grid style={draw=vpgray!18},
        nodes near coords,
        point meta=explicit symbolic,
        every node near coord/.append style={
            font=\scriptsize,
            anchor=south,
            yshift=3pt,
            inner sep=0.5pt,
            text=black
        },
        legend style={
            at={(0.5,-0.18)},
            anchor=north,
            legend columns=3,
            font=\scriptsize,
            draw=none,
            fill=none,
            row sep=1pt
        },
        legend cell align={left}
    ]
    \addplot[
        fill=vpgray!45,
        draw=vpgray!80
    ] coordinates {
        (Mug,60) [60]
        (Card,11) [11]
        (Laptop,36) [36]
        (Cup,17) [17]
        (Cabinet,63) [63]
        (Avg.,37.4) [{\makebox[9pt][c]{37}}]
    };
    \addplot[
        fill=vporange!65,
        draw=vporange!95
    ] coordinates {
        (Mug,62) [62]
        (Card,33) [33]
        (Laptop,18) [18]
        (Cup,20) [20]
        (Cabinet,97) [97]
        (Avg.,46.0) [{\makebox[9pt][c]{46}}]
    };
    \addplot[
        fill=vpgreen!70,
        draw=vpgreen!95
    ] coordinates {
        (Mug,89) [89]
        (Card,64) [64]
        (Laptop,89) [89]
        (Cup,52) [52]
        (Cabinet,98) [98]
        (Avg.,78.4) [{\makebox[9pt][c]{78}}]
    };
    \legend{
        Baseline,
        Random-state Object+EE,
        Interaction-aware Object+EE
    }
    \end{axis}
    \end{tikzpicture}%
  }
  \caption{Unseen-view success rates for the baseline and Object+EE
  augmentation with random or interaction-aware states.}
  \label{fig:interaction_state_bar}
\end{figure}
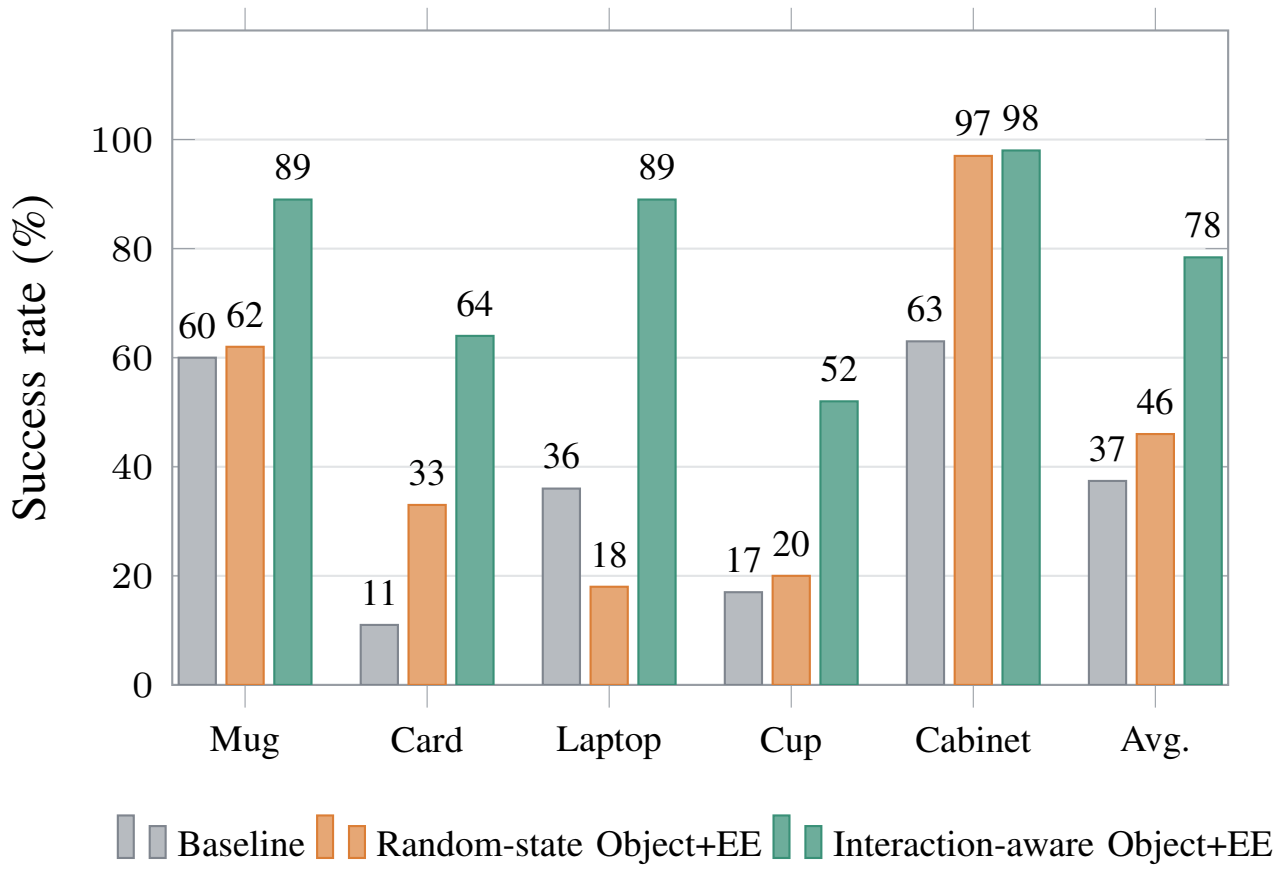

Our method substantially and consistently improves the success rate across tasks. In contrast, random-state augmentation is unstable and can even perform worse than the baseline on some tasks, demonstrating the importance of interaction-aware state sampling and joint object--end-effector supervision.

\subsection{Generalization to OOD Variations}

Finally, we study whether the benefit extends beyond camera-extrinsic changes to object, scene, and camera-intrinsic distribution shifts. We compare the baseline with our method.

\begin{table*}[!t]
    \centering
    \caption{\normalfont Closed-loop success rates under OOD variations. The evaluation protocol specifies 20 episodes per OOD category.}
    \label{tab:ood_variations}
    \small
    \setlength{\tabcolsep}{5pt}
    \begin{tabular*}{\textwidth}{@{\extracolsep{\fill}}lccc@{}}
        \toprule
        Task & \shortstack{Object OOD\\(Baseline SR / our method SR)} & \shortstack{Scene OOD\\(Baseline SR / our method SR)} & \shortstack{Camera-Intrinsic OOD\\(Baseline SR / our method SR)} \\
        \midrule
        Hanging Mug & 20\% / \textbf{50\%} & 20\% / \textbf{50\%} & 70\% / \textbf{80\%} \\
        Move Playing Card Away & 5\% / \textbf{25\%} & 5\% / \textbf{10\%} & 35\% / \textbf{45\%} \\
        Open Laptop & \textbf{95\%} / 90\% & 0\% / 0\% & 40\% / 40\% \\
        Place Empty Cup & 0\% / \textbf{5\%} & 0\% / 0\% & 0\% / \textbf{5\%} \\
        Put Object in Cabinet & 0\% / 0\% & 45\% / \textbf{65\%} & 55\% / \textbf{95\%} \\
        \textbf{Five-task mean SR} & 24.0\% / \textbf{34.0\%} & 14.0\% / \textbf{25.0\%} & 40.0\% / \textbf{53.0\%} \\
        \bottomrule
    \end{tabular*}
\end{table*}

Overall, our method provides modest improvements in mean SR under all three OOD conditions, but the gains do not extend to every task. This may be due to the distribution gap between the object-augmentation data and the expert demonstrations, together with the limited capacity of the policy model.

\section{REAL-WORLD EXPERIMENTS}
\label{sec:real_world_experiments}

We conduct real-robot experiments to answer the following questions: (1) Can
geometric data augmentation improve generalization to unseen viewpoints when
deployed on a physical robot in real-world scenes? (2)
Without reconstructing the physical task environment as a digital twin in
simulation, does simulation-derived geometric supervision transfer more
effectively than simulation-derived action supervision under viewpoint shifts?
(3) What are the respective roles of simulation-derived and real-image-derived
geometric supervision in improving policy performance across viewpoint shifts,
and are their effects complementary? (4) Does geometric co-training encourage
the learned policy representation to focus on task-relevant image regions?

\begin{table*}[!t]
    \centering
    \caption{\normalfont Aggregate real-robot successes across four unseen
    camera viewpoints. Each task reports successes out of 40 trials (10 per
    view), and Overall reports successes out of 160 trials. Best results are
    bold.}
    \label{tab:real_robot_unseen_views}
    \small
    \setlength{\tabcolsep}{5pt}
    \renewcommand{\arraystretch}{1.05}
    \begin{tabular*}{\textwidth}{@{\extracolsep{\fill}}lrrrrr@{}}
        \toprule
        Method & \shortstack{Place Empty\\Cup} & \shortstack{Hanging\\Mug} &
        \shortstack{Open\\Laptop} & \shortstack{Pull\\Drawer} & Overall \\
        \midrule
        Baseline & 3/40 & 0/40 & 1/40 & 1/40 & 5/160 \\
        Sim Action Aug. & 7/40 & 15/40 & 6/40 & 0/40 & 28/160 \\
        Sim Pose Aug. & 9/40 & 11/40 & 26/40 & 0/40 & 46/160 \\
        Real Pose Aug. & 12/40 & 19/40 & 12/40 & 2/40 & 45/160 \\
        VISTA (DROID) & 15/40 & 11/40 & 8/40 & 15/40 & 49/160 \\
        Real+Sim Pose Aug. (\method{}, Ours) & \textbf{27/40} &
        \textbf{28/40} & \textbf{32/40} & \textbf{22/40} &
        \textbf{109/160} \\
        \bottomrule
    \end{tabular*}
\end{table*}

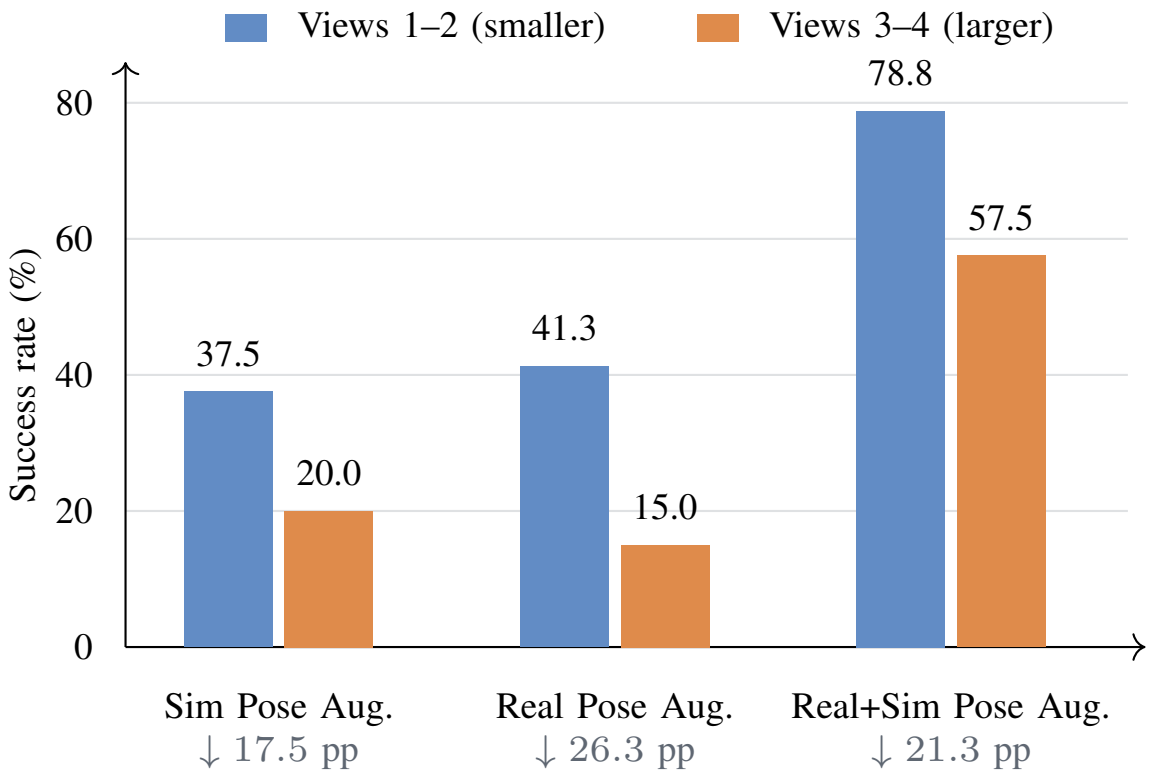
\begin{figure}[!t]
    \centering
    \begin{tikzpicture}[x=0.78cm,y=0.045cm,font=\scriptsize]
        \foreach \y in {0,20,40,60,80} {
            \draw[vpgray!20] (0.8,\y) -- (9.3,\y);
            \node[anchor=east] at (0.68,\y) {\y};
        }
        \draw[->] (0.8,0) -- (9.45,0);
        \draw[->] (0.8,0) -- (0.8,86);
        \node[rotate=90] at (-0.05,40) {Success rate (\%)};

        \fill[vpblue!80] (1.65,89) rectangle (2.0,93);
        \node[anchor=west] at (2.1,91) {Views 1--2 (smaller)};
        \fill[vporange!85] (5.65,89) rectangle (6.0,93);
        \node[anchor=west] at (6.1,91) {Views 3--4 (larger)};

        \fill[vpblue!80] (1.3,0) rectangle (2.05,37.5);
        \fill[vporange!85] (2.15,0) rectangle (2.9,20.0);
        \fill[vpblue!80] (4.15,0) rectangle (4.9,41.3);
        \fill[vporange!85] (5.0,0) rectangle (5.75,15.0);
        \fill[vpblue!80] (7.0,0) rectangle (7.75,78.8);
        \fill[vporange!85] (7.85,0) rectangle (8.6,57.5);

        \node[anchor=south] at (1.675,38.5) {37.5};
        \node[anchor=south] at (2.525,21.0) {20.0};
        \node[anchor=south] at (4.525,42.3) {41.3};
        \node[anchor=south] at (5.375,16.0) {15.0};
        \node[anchor=south] at (7.375,79.8) {78.8};
        \node[anchor=south] at (8.225,58.5) {57.5};

        \node[anchor=north,align=center] at (2.1,-4)
        {Sim Pose Aug.\\\textcolor{vpgray}{$\downarrow 17.5$ pp}};
        \node[anchor=north,align=center] at (4.95,-4)
        {Real Pose Aug.\\\textcolor{vpgray}{$\downarrow 26.3$ pp}};
        \node[anchor=north,align=center] at (7.8,-4)
        {Real+Sim Pose Aug.\\\textcolor{vpgray}{$\downarrow 21.3$ pp}};
    \end{tikzpicture}
    \caption{\normalfont Real-robot success rates under smaller (Views 1--2)
    and larger (Views 3--4) viewpoint shifts for each geometric-supervision
    source. Each bar aggregates 80 trials; annotations show the decrease from
    smaller to larger shifts.}
    \label{fig:real_pose_source_roles}
\end{figure}

\subsection{Experimental Setup}

\subsubsection{Task}

We evaluate on an AgileX ALOHA robot using four manipulation tasks:
\textbf{Place Empty Cup}, \textbf{Hanging Mug}, \textbf{Open Laptop}, and
\textbf{Pull Drawer}. The four tasks respectively evaluate pick-and-place,
precise insertion and hanging, interaction with a revolute articulation, and
interaction with a prismatic articulation.
\subsubsection{Data}

For each task, we collect 100 successful real-robot demonstrations. Each
demonstration contains RGB observations from a fixed external
camera and a wrist camera. We augment these demonstrations using the two pipelines
described in Secs.~\ref{sec:simulation_augmentation}
and~\ref{sec:real_world_augmentation}. For each task, the simulation pipeline
generates 100,000 pose-labeled samples for the corresponding manipulated object.
The real-world pipeline annotates every external-camera frame from all 100
demonstrations, and these annotated frames constitute the real pose dataset.
During co-training, we sample one action mini-batch and one geometric mini-batch
at each optimization step.

\subsubsection{Evaluation}

All methods use the same Action Chunking Transformer (ACT) backbone. For each task, we evaluate four external-camera viewpoints that are not observed during demonstration collection. Views 1 and 2
have smaller camera-pose offsets from the data-collection view, whereas Views 3
and 4 have larger offsets. Each method is tested for 10 trials at every task--view pair.

\begin{figure*}[t]
    \centering
    \begingroup
    \setlength{\tabcolsep}{0.8pt}
    \setlength{\fboxsep}{0.25pt}
    \setlength{\fboxrule}{0.6pt}
    \newcommand{\realtrainview}[1]{%
        \raisebox{-0.5\height}{\fcolorbox{vpblue}{white}{\includegraphics[width=0.155\textwidth]{#1}}}}
    \newcommand{\realsmallview}[1]{%
        \raisebox{-0.5\height}{\fcolorbox{vpgreen}{white}{\includegraphics[width=0.155\textwidth]{#1}}}}
    \newcommand{\reallargeview}[1]{%
        \raisebox{-0.5\height}{\fcolorbox{vporange}{white}{\includegraphics[width=0.155\textwidth]{#1}}}}
    \scriptsize
    \begin{tabular}{@{}c@{\hspace{1.5pt}}ccccc@{}}
        & \multicolumn{1}{c}{\textcolor{vpblue}{\textbf{Data-collection view}}}
        & \multicolumn{2}{c}{\textcolor{vpgreen}{\textbf{Smaller viewpoint shifts}}}
        & \multicolumn{2}{c}{\textcolor{vporange}{\textbf{Larger viewpoint shifts}}} \\[-0.2mm]
        \textbf{Task} & \textbf{View $c_0$} & \textbf{View 1} & \textbf{View 2} & \textbf{View 3} & \textbf{View 4} \\[0.35mm]

        \textbf{\shortstack{Place Empty\\Cup}}
        & \realtrainview{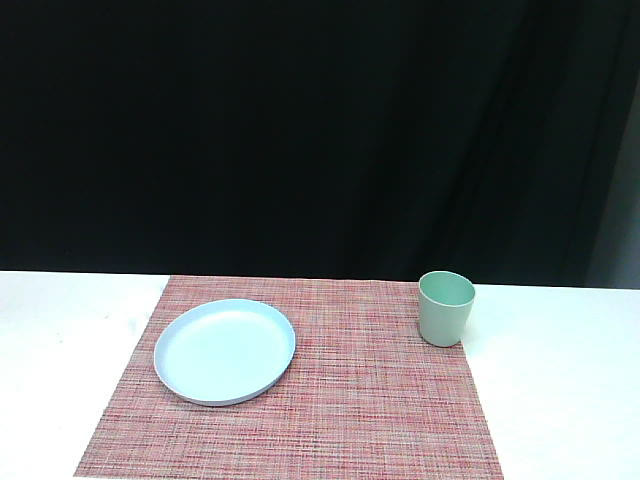}
        & \realsmallview{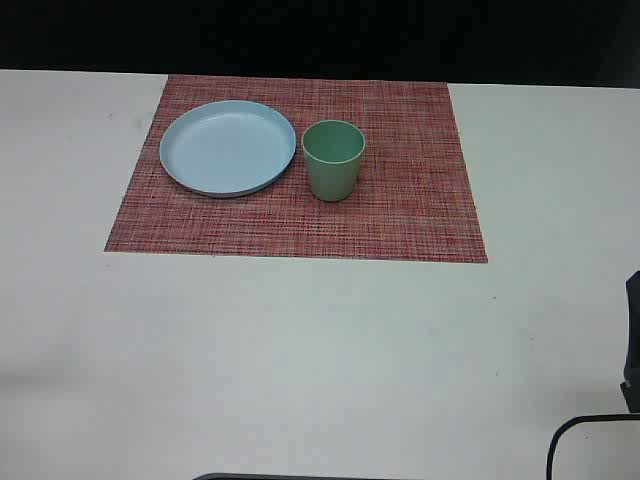}
        & \realsmallview{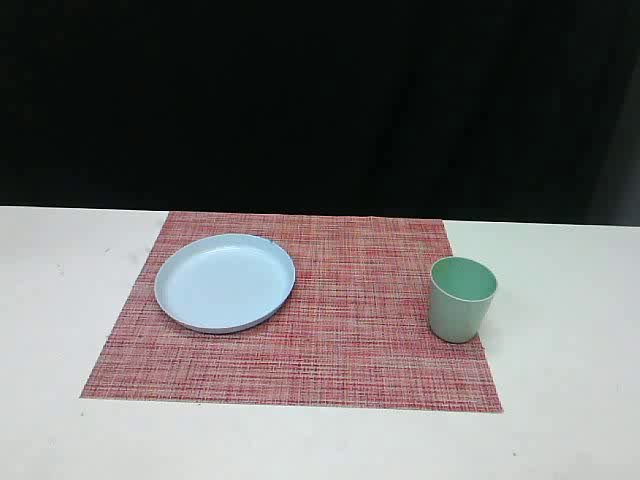}
        & \reallargeview{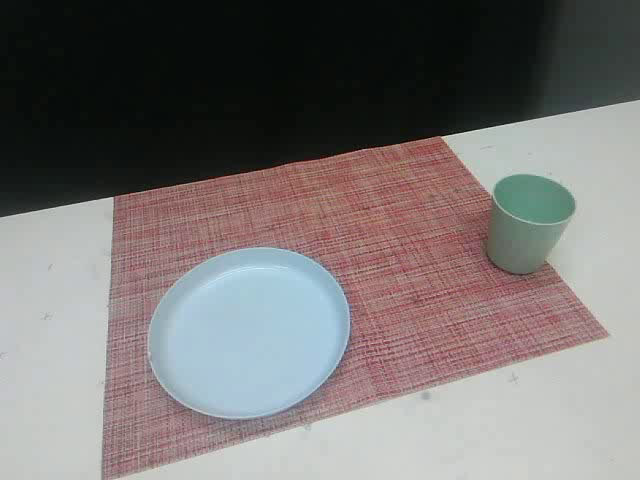}
        & \reallargeview{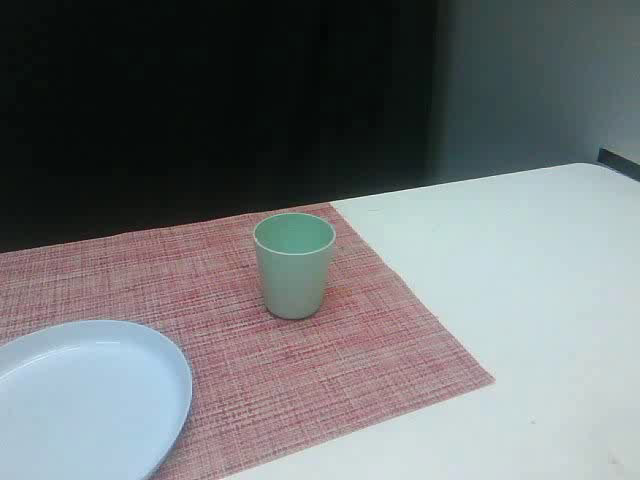} \\[0.35mm]

        \textbf{\shortstack{Hanging\\Mug}}
        & \realtrainview{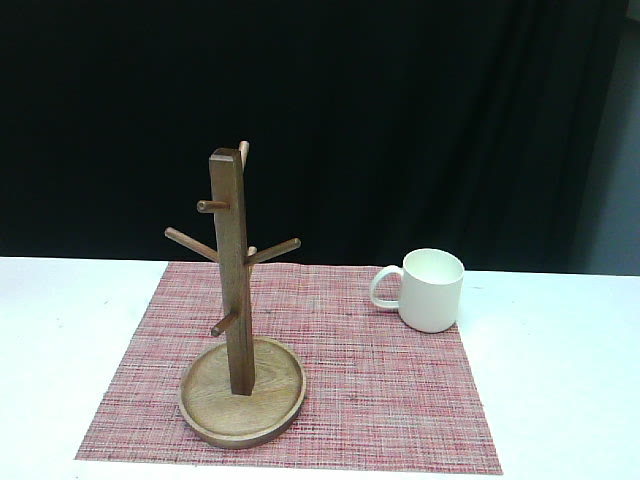}
        & \realsmallview{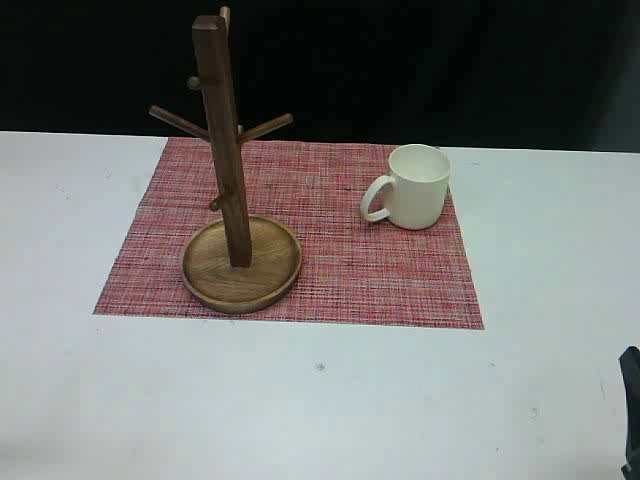}
        & \realsmallview{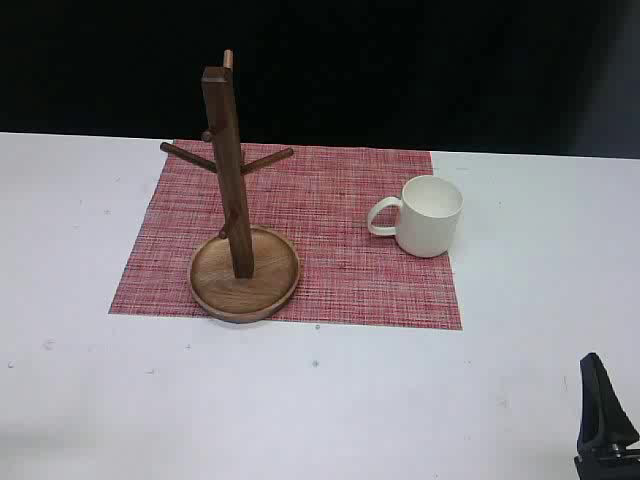}
        & \reallargeview{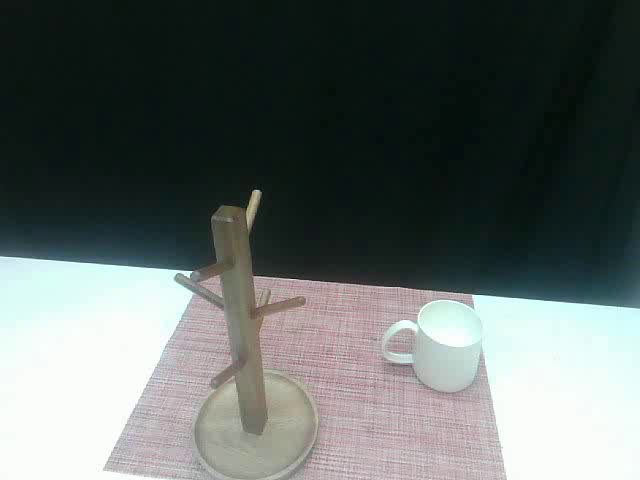}
        & \reallargeview{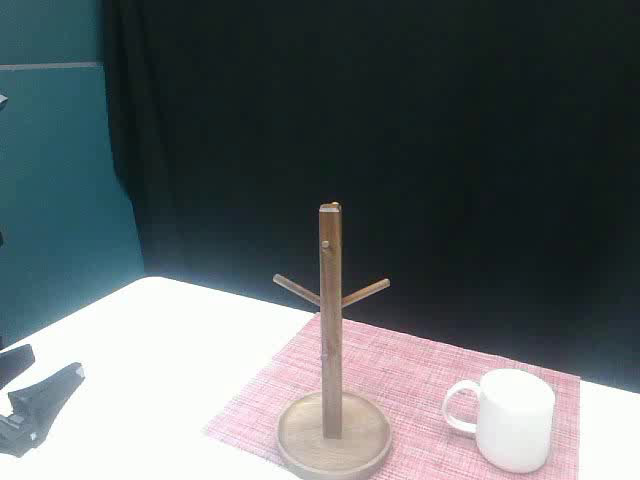} \\[0.35mm]

        \textbf{\shortstack{Open Laptop}}
        & \realtrainview{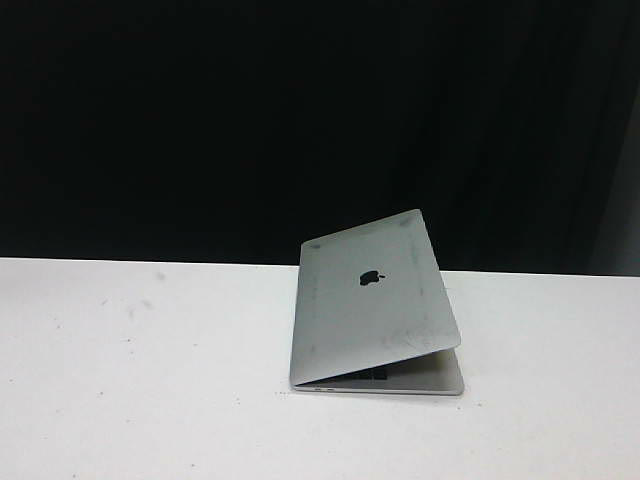}
        & \realsmallview{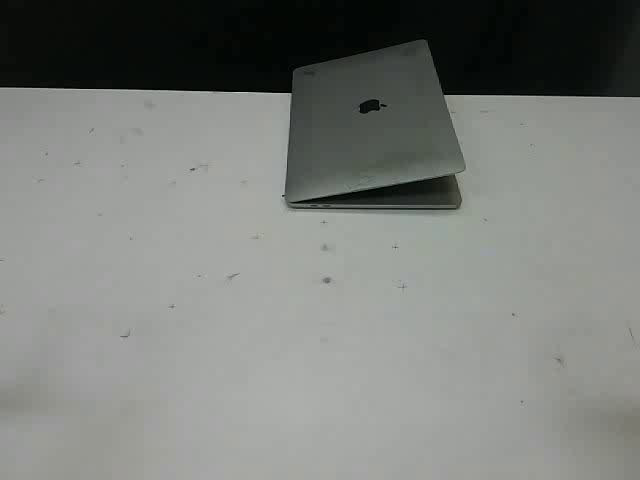}
        & \realsmallview{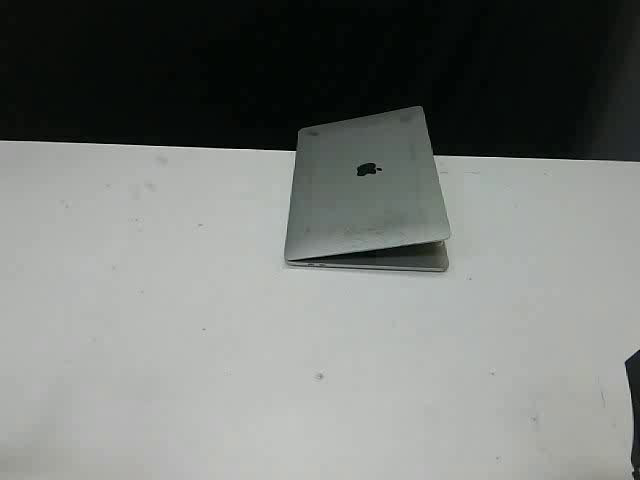}
        & \reallargeview{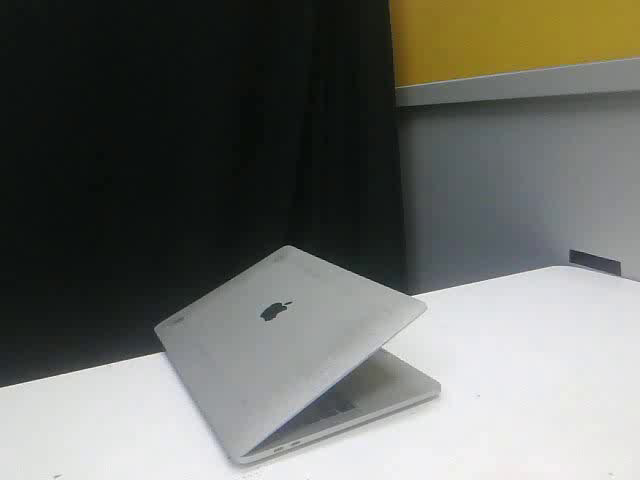}
        & \reallargeview{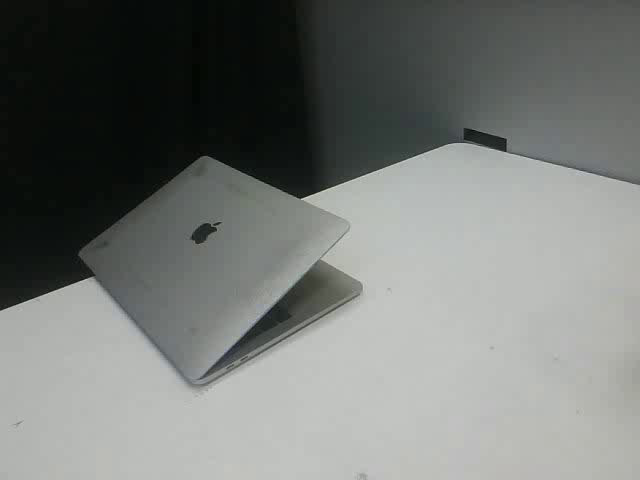} \\[0.35mm]

        \textbf{\shortstack{Pull Drawer}}
        & \realtrainview{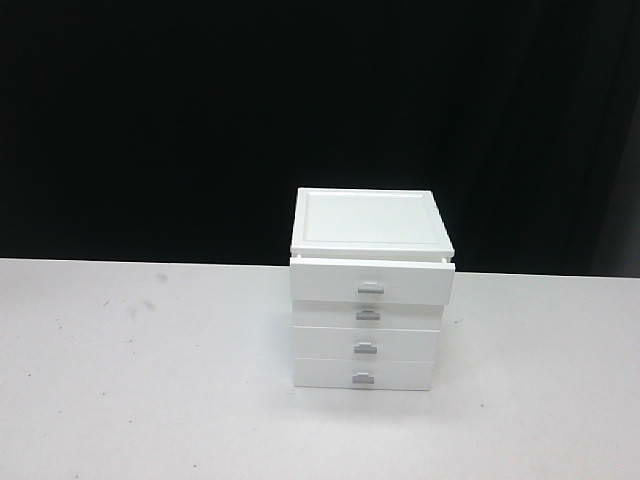}
        & \realsmallview{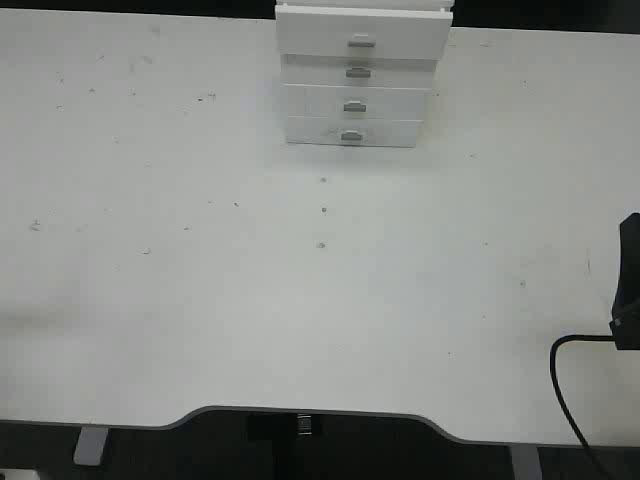}
        & \realsmallview{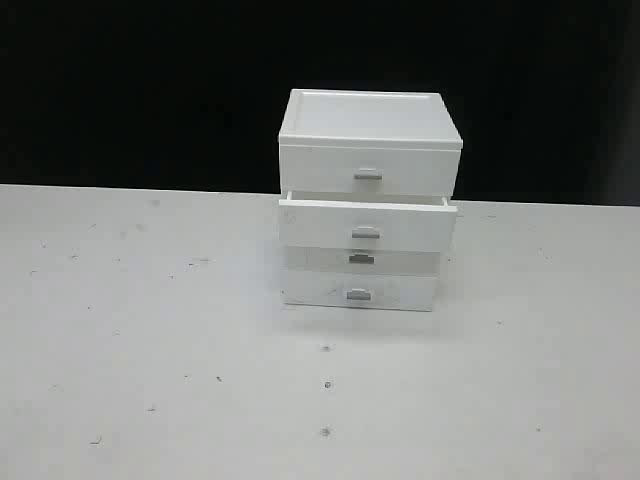}
        & \reallargeview{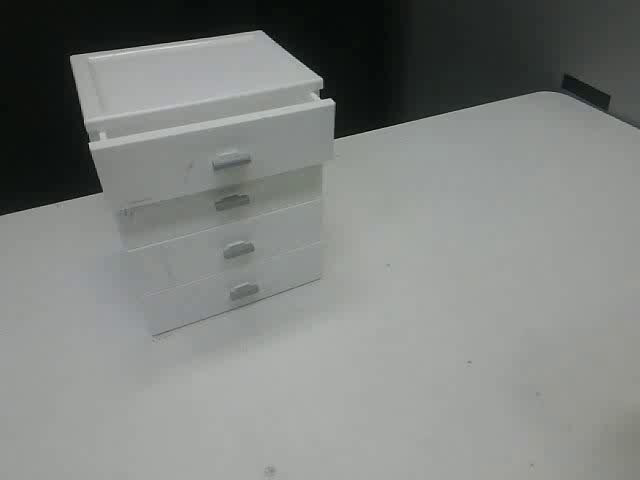}
        & \reallargeview{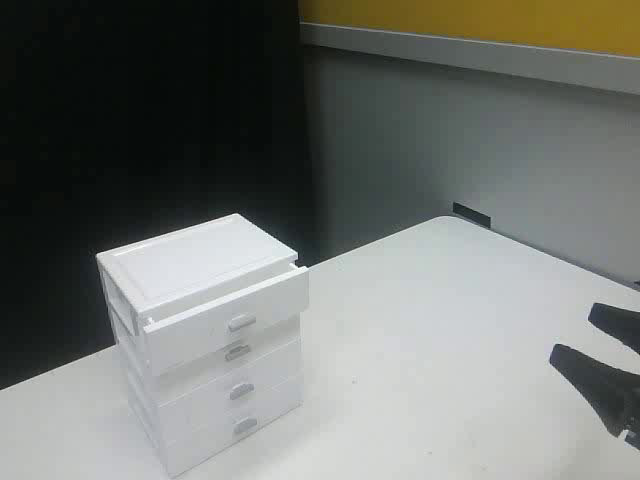}
    \end{tabular}
    \endgroup
    \caption{\normalfont Real-world camera configurations used for data
    collection and evaluation. Each row shows one task. Blue denotes the fixed
    external-camera view used for demonstration collection; green and orange
    denote unseen evaluation views with smaller (Views 1--2) and larger
    (Views 3--4) camera-pose offsets, respectively. Evaluation images are the
    external-camera first frames extracted from the corresponding rollout
    videos.}
    \label{fig:real_evaluation_views}
\end{figure*}

\begin{figure*}[!t]
    \centering
    \begingroup
    \setlength{\tabcolsep}{1.25pt}
    \newcommand{\attentiontile}[1]{%
        \includegraphics[width=0.190\textwidth]{#1}}
    \scriptsize
    \begin{tabular}{@{}r@{\hspace{1.5pt}}cccc@{}}
        & \textbf{Place Empty Cup}
        & \textbf{Hanging Mug}
        & \textbf{Open Laptop}
        & \textbf{Pull Drawer} \\[0.35mm]
        \raisebox{0.37in}{\textbf{Baseline}}
        & \attentiontile{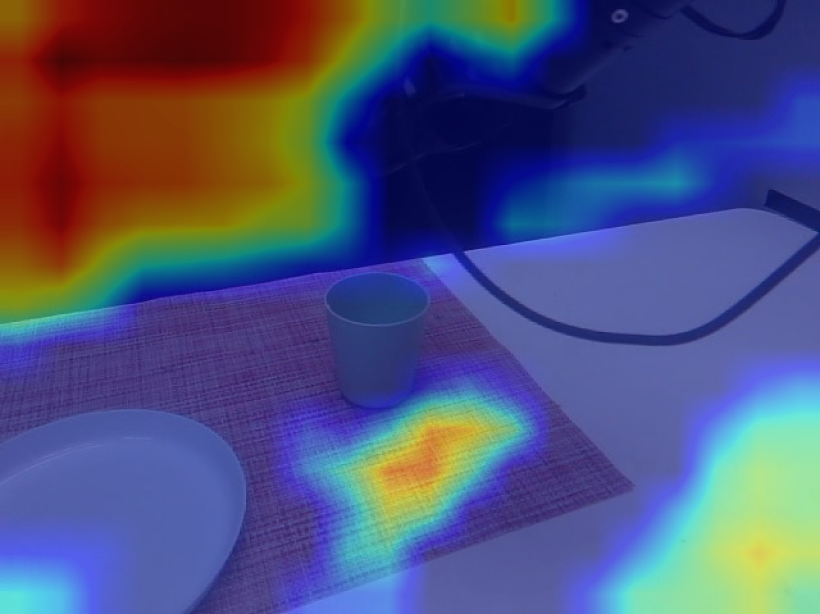}
        & \attentiontile{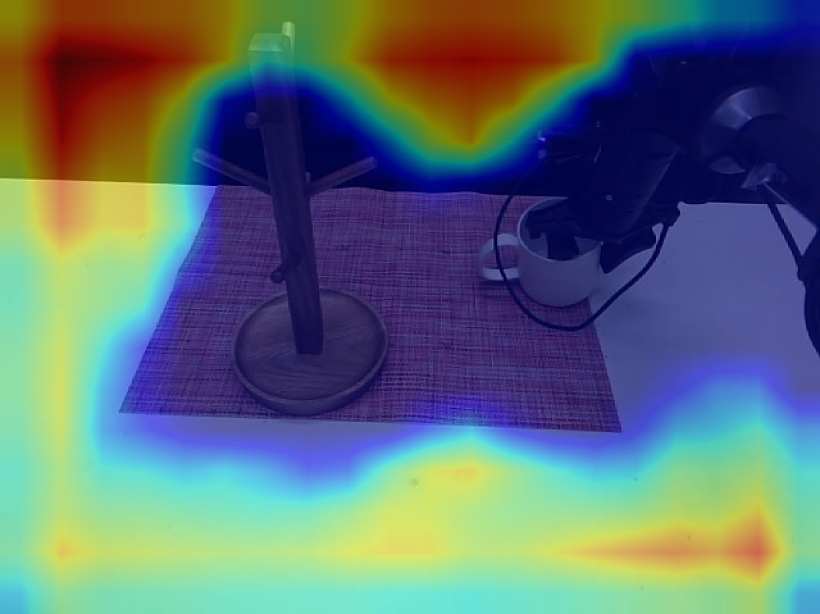}
        & \attentiontile{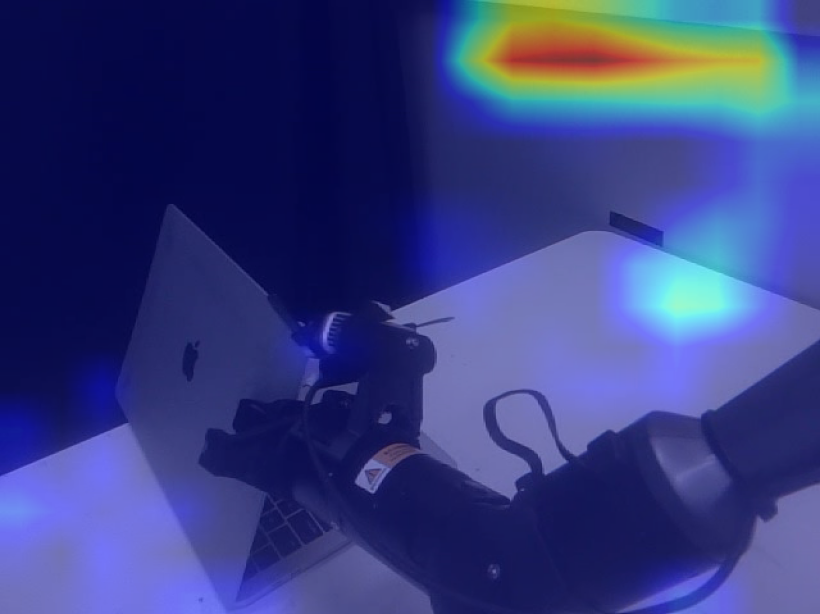}
        & \attentiontile{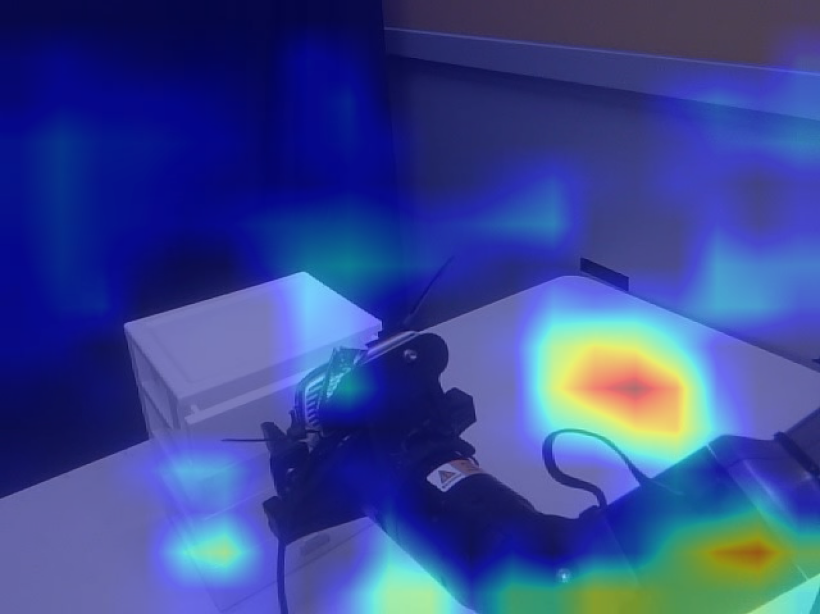} \\[0.35mm]
        \raisebox{0.37in}{\textbf{\method{}}}
        & \attentiontile{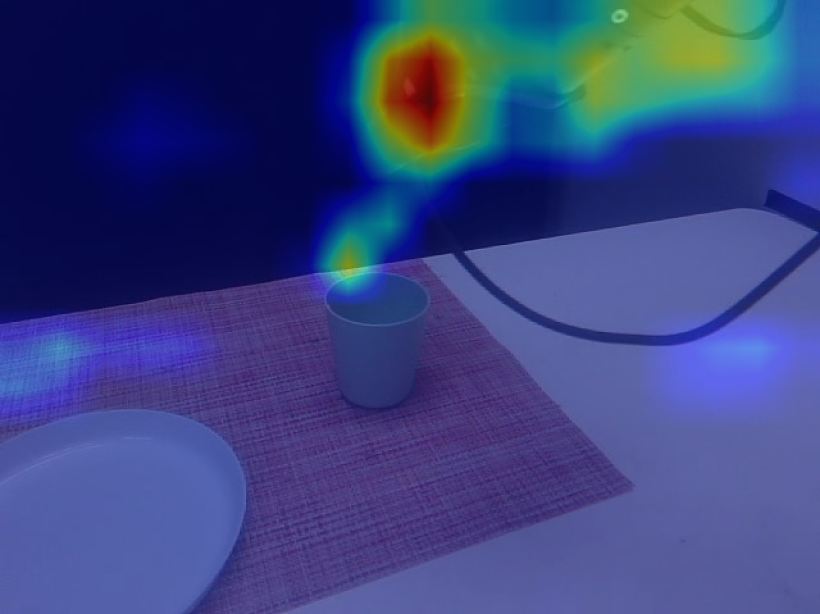}
        & \attentiontile{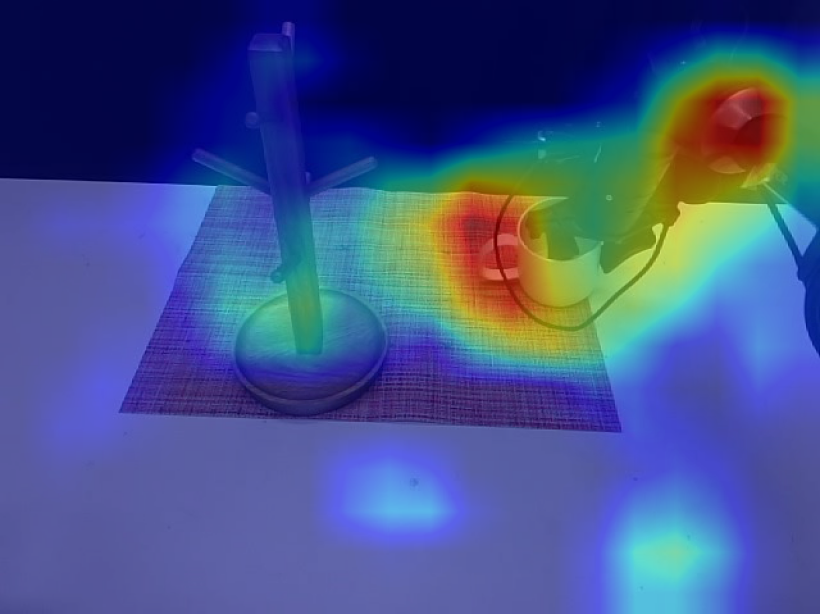}
        & \attentiontile{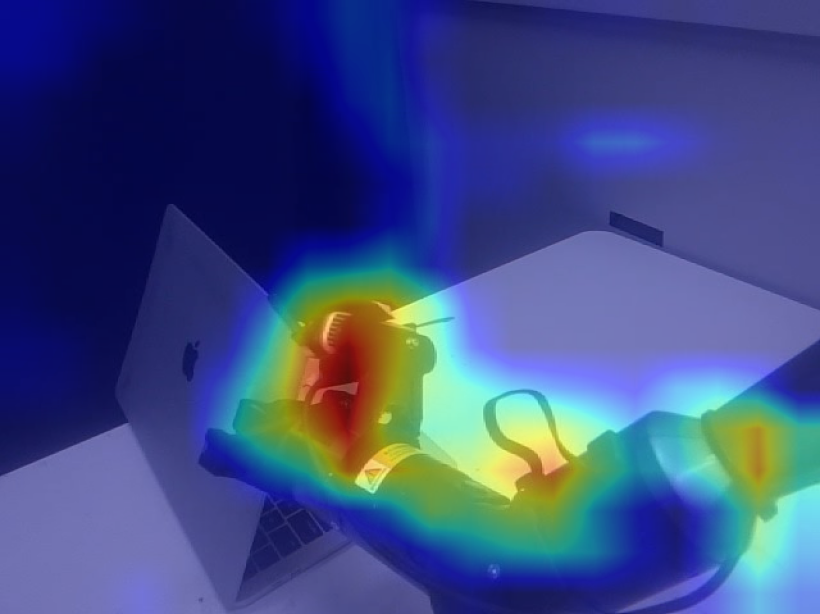}
        & \attentiontile{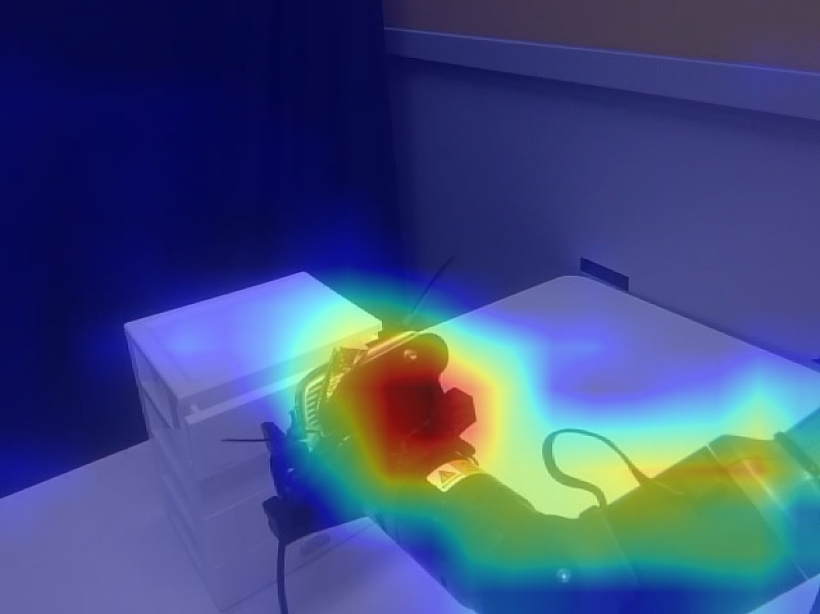}
    \end{tabular}
    \endgroup
    \caption{\normalfont Grad-CAM responses computed over action queries 0--9
    for the baseline and \method{} on matched real-robot frames from all four
    tasks. Warmer colors indicate higher responses. The baseline often responds
    to background regions, whereas \method{} concentrates more strongly on the
    manipulated object, end effector, and local interaction region.}
    \label{fig:real_attention_maps}
\end{figure*}

\subsubsection{Compared Methods}

We compare the following six variants.

\begin{itemize}
    \item \textbf{Baseline} is trained only on the 100 real action
    demonstrations collected for each task.

    \item \textbf{Sim Action Aug.} uses the motion-planned trajectories
    from the simulation augmentation pipeline as additional action-supervised
    data.

    \item \textbf{Sim Pose Aug.} combines the same real action demonstrations
    with simulated pose labels.

    \item \textbf{Real Pose Aug.} combines the same real action demonstrations
    with real-image object-pose labels.
    \item \textbf{VISTA (DROID)} augments the real demonstrations with novel
    external-camera views synthesized using the DROID-finetuned ZeroNVS model,
    following VISTA~\cite{tian2025vista}. The synthesized external views retain
    the action labels of their source demonstrations, while the wrist-camera
    observations remain unchanged.

    \item \textbf{Real+Sim Pose Aug. (\method{}, Ours)} combines the real action
    demonstrations with simulated object- and end-effector-pose supervision and
    real object-pose supervision.
\end{itemize}

\subsection{Results}

\noindent\textbf{Q1: Does geometric augmentation improve real-world
manipulation?}
As shown in Table~\ref{tab:real_robot_unseen_views}, both sources of geometric
supervision substantially improve closed-loop performance under unseen
viewpoints. Sim Pose Aug. and Real Pose Aug. raise the overall mean success rate
from $3.1\%$ for the baseline to $28.8\%$ and $28.1\%$, respectively. Combining
the two sources achieves the highest overall mean of $68.1\%$ and the best mean
on every task.

\noindent\textbf{Q2: Without a digital twin, does simulation transfer better
through geometry than actions?}
In this non-digital-twin setting, Sim Pose Aug. achieves a higher overall mean
than Sim Action Aug. ($28.8\%$ vs. $17.5\%$). Although the gains vary by task,
the average improvement indicates that simulation transfers more effectively
through geometric supervision than through action labels. Qualitatively, when
the sim-to-real object gap is small, Sim Action Aug. can transfer local
primitives such as grasping but often fails at later interaction stages that
are difficult to reproduce in simulation, preventing completion of the full
task.

\noindent\textbf{Q3: What are the respective roles of simulation-derived and
real-image-derived geometric supervision, and are their effects complementary?}
Despite similar overall means ($28.8\%$ and $28.1\%$), the two sources exhibit
different sensitivity to camera shift. Averaged over tasks, Sim Pose Aug.
achieves $37.5\%$ on the smaller shifts (Views 1--2) and $20.0\%$ on the larger
shifts (Views 3--4), a drop of $17.5$ percentage points. Real Pose Aug. is
stronger on the smaller shifts ($41.3\%$) but falls to $15.0\%$ on the larger
shifts, a larger drop of $26.3$ points. This trend suggests that
simulation-derived supervision
contributes broader viewpoint coverage, whereas real-image-derived supervision
primarily strengthens grounding in the real scene near the data-collection
view. Notably, the two sources are strongly complementary: Real+Sim Pose Aug.
reaches $78.8\%$ on Views 1--2 and retains $57.5\%$ on Views 3--4.

\noindent\textbf{Q4: Does geometric co-training focus the representation on
task-relevant regions?}
Figure~\ref{fig:real_attention_maps} compares Grad-CAM responses computed over
action queries 0--9 for the baseline and \method{} on matched real-robot frames
from camera viewpoints unseen during training. Across all four tasks, the
baseline often activates on background surfaces or other off-task regions,
whereas \method{} concentrates more strongly on the manipulated object, end
effector, and their interaction region. This qualitative evidence suggests that
pose supervision shapes the shared visual representation toward geometry that
is relevant for action prediction.

\FloatBarrier
\nocite{wu2026galatea}
\bibliographystyle{IEEEtran}
\bibliography{references}

\end{document}